\documentclass[11pt]{article}
\usepackage{eamt26}
\usepackage{times}
\usepackage{url}
\usepackage{latexsym}
\usepackage[small,bf]{caption} 
\usepackage{booktabs}
\usepackage{threeparttable}
\usepackage[dvipsnames]{xcolor}
\usepackage{soul}
\usepackage{amsmath}
\sethlcolor{yellow!40}

\newcommand{\confname}{EAMT 2026}

\newcommand{\researchtechlength}{$10$ (ten)}
\newcommand{\researchtransuserslength}{$10$ (ten)}
\newcommand{\implementationslength}{$6$ (six)}
\newcommand{\projectlength}{$2$ (two)}

\title{Smarter edits? \\Post-editing with error highlights and translation suggestions}

\author{Fleur V.J. van Tellingen\textsuperscript{1}, Gautam Ranka\textsuperscript{2}, Dora Žugčić \textsuperscript{3}, Joyce van der Wal\textsuperscript{4}, \\ \textbf{Andrea	Camasta\textsuperscript{5}, Livio	Guerra\textsuperscript{5}, Alina Karakanta\textsuperscript{1}} \\
  \textsuperscript{1}Leiden University Centre for Linguistics \\
  \textsuperscript{2}Visvesvaraya National Institute of Technology \\ 
  \textsuperscript{3}Department of Bionanoscience, Faculty of Applied Sciences, Delft University of Technology \\
  \textsuperscript{4}Pedagogical Sciences, Leiden University\\ \textsuperscript{5}Faculty of Science, Leiden University \\
  \texttt{a.karakanta@leidenuniv.nl}
  }

\date{}

\begin{document}
\maketitle
\begin{abstract}
  As MT quality increases, interest in enhanced post-editing features such as QE-derived error highlights is growing, yet evidence for their usefulness remains limited. In this work, we explore the usefulness of LLM-derived error highlights and correction suggestions based on automatic post-editing (APE). We conduct a study where professional translators (En$\rightarrow$Nl) post-edit translations using APE error highlights and correction suggestions and compare productivity, quality and user experience to regular PE and PE with QE-derived highlights. While no condition yielded productivity or quality gains compared to regular PE, APE highlights were better received than QE-derived highlights, and correction suggestions improved the overall user experience. 
\end{abstract}

\section{Introduction}

Generative large language models (LLMs) have demonstrated remarkable proficiency not only for machine translation (MT) but across several translation-related tasks, among which assessing MT quality \cite{kocmi-federmann-2023-large}, and fine-grained error detection and correction \cite{fernandes-etal-2023-devil,lu-etal-2024-error}. Despite the growing interest in explainable MT evaluation, the usefulness of LLM suggestions in augmenting translators' workflows has been scarcely explored. Previous work has focused heavily on quality estimation (QE) features for enhancing translators’ productivity, but the results were inconclusive, with 
limited or no productivity gains \cite{bechara-21MTQEPE,shenoy-etal-2021-investigating,teixeira-obrien-2017-impact}
, even when errors were derived from human edits rather than QE \cite{sarti2025qe4pewordlevelqualityestimation}. 
Other studies have indicated that providing more intelligible feedback, such as explicit translation suggestions, can support translators more effectively \cite{10.1145/3173574.3174098}.

In this work, we investigate whether signals derived from automatic post-editing (APE) instead of QE can enable smarter editing decisions and enhance translators’ productivity, experience and confidence. As shown in Table \ref{tab:conditions}, error highlights obtained from QE ({\color{orange}H-QE}) sometimes only cover parts of words. In addition, there is no justification why highlighted characters are considered an error. Correction suggestions ({\color{SeaGreen}S-APE}) give translators more interpretable guidance about where and how to intervene. 
We thus ask the following questions:
Could APE suggestions enhance \textbf{productivity} by helping translators edit machine-translated texts more efficiently? 
As MT outputs improve in quality, could such features assist translators in detecting errors, resulting in \textbf{higher-quality} translations? What is their impact on translators' \textbf{perception and confidence}?

\begin{table}[]
    \centering
    \begin{footnotesize}
    \begin{tabular}{l|l} \toprule
SRC & \begin{tabular}[l]{@{}l@{}}
          How to find out if you're flying on a Boeing?\end{tabular}\\
REF & Hoe komt u erachter of u in een Boeing vliegt?\\
\midrule
\color{blue}{PE}      & 
Hoe kunt u ontdekken of u vliegt in een Boeing?
                      \\
\color{orange}{H-QE}    & Hoe kunt u \hl{ontdekke}n of u vliegt \hl{in} een Boeing? \\
\color{magenta}{H-APE}  & Hoe kunt u \hl{ontdekken} of u vliegt in een Boeing? \\
\color{SeaGreen}{S-APE} & \begin{tabular}[l]{@{}l@{}}
Hoe kunt u \hl{ontdekken}\colorbox{green}{(erachter komen)} of u \\ vliegt in een Boeing?                         \end{tabular} \\ \bottomrule
    \end{tabular}
    \end{footnotesize}
    \caption{Example of highlights derived by QE ({\color{orange}H-QE}) and APE ({\color{magenta}H-APE}), together with correction suggestions ({\color{SeaGreen}S-APE}: text in parentheses). }
    \label{tab:conditions}
\end{table}

To answer these questions, we conducted a user study incorporating automatic error annotations and correction suggestions in a realistic post-editing setting. Eight English$\rightarrow$Dutch professional translators post-edited machine translated texts in two domains (news, biomedical) in four conditions: 
1) simple post-editing ({\color{blue}PE}), 2) post-editing with error highlights from quality estimation ({\color{orange}H-QE}), 3) post-editing with error highlights from automatic post-editing ({\color{magenta}H-APE}) and 4) post-editing with error highlights and correction suggestions ({\color{SeaGreen}S-APE}). The PE task was conducted in SmartPE, an interface incorporating LLM suggestions, which logs granular editing data. We collected product and 
process data, 
and user ratings on perceived quality of MT, error highlights and suggestions
. Lastly, translators provided qualitative feedback through semi-structured post-task interviews. Our contributions are as follows:


\begin{itemize}
    \item A rich multi-parallel post-editing dataset (14.400 words) combining raw MT with automatic error annotations and correction suggestions, and 8 PE versions, all of them evaluated following the ESA protocol~\cite{kocmi-etal-2024-error}, 
    keystrokes, 
    user perception scores and quality ratings in two domains.\footnote{Data and code: \url{https://github.com/fatalinha/smarter-edits}}
    \item \textbf{SmartPE}: A new post-editing interface  incorporating error highlights and translation suggestions, released open-source.\footnote{\url{https://github.com/fatalinha/SmartPE}}
    \item An analysis of productivity, quality, and user experience in post-editing with various degrees of LLM assistance.
\end{itemize}

We publicly release the data, code and post-editing interface to facilitate future work on LLM suggestions in post-editing workflows.

\section{Related work}
\subsection{Explainable MT assessment}
Automatic MT evaluation has shifted from string overlap metrics~\cite{papineni-etal-2002-bleu,popovic-2015-chrf}, which yield a single holistic score, 
toward learned metrics and fine-grained QE systems providing insight into the nature or location of translation errors
~\cite{kepler-etal-2019-openkiwi}
. 
For instance, \mbox{xCOMET}~\cite{guerreiro2023xcomettransparentmachinetranslation} unifies segment-level quality prediction with span detection, identifying both the location and severity of errors within a single model. Concurrently, the LLM-as-judge paradigm~\cite{zheng-etal-23-judges} has been applied to MT evaluation through tools such as Auto-MQM \cite{fernandes-etal-2023-devil}, EAPrompt \cite{lu-etal-2024-error} and GEMBA-MQM~\cite{kocmi-federmann-2023-gemba,kocmi-etal-2024-error}, which prompt LLMs with quality criteria to elicit span-level annotations aligned with the MQM error typology~\cite{mqm-2014} or free-text explanations~\cite{treviso2024xtower}.

Building on QE signals, a natural extension is to move from error detection to error correction through APE. In this paired paradigm, APE systems leverage QE predictions to guide targeted corrections for raw MT output~\cite{chatterjee-etal-2018-combining,fernandes-etal-2023-devil,deoghare-etal-2023-quality,treviso2024xtower}.
Recently, LLMs were shown to produce meaningful and targeted edits that improve overall translation quality~\cite{raunak-etal-2023-leveraging,briakou-etal-2024-translating}. Nevertheless, LLMs show a different PE behaviour than professional translators, missing errors and producing hallucinated edits, or performing preferential edits which translators are instructed to avoid~\cite{macken-2024-machine,deoghare-etal-2025-giving}. Despite this, multi-agent frameworks that simulate collaborative practices in human translation workflows are increasingly used in industry~\cite{briva-iglesias-2025-ai,wu-etal-2025-perhaps}.
In this work, we leverage QE paired with APE signals to generate PE suggestions in translators' workflows.

\subsection{Post-editing with quality estimation}
The potential of QE to enhance post-editing productivity has been explored in a number of user studies, with mixed and often inconclusive results. Earlier studies focused on sentence-level QE in statistical MT, by displaying a quality score or using a traffic light systems (e.g. red, yellow, green) suggesting the amount of PE required. These approaches found QE to improve PE efficiency only under specific conditions, such as when MT output quality 
was low~\cite{bechara-21MTQEPE} or when QE predictions were sufficiently accurate~\cite{escartin2017questing,turchi-etal-2015-mt}. 
\newcite{teixeira-obrien-2017-impact} found no significant effects on either technical or cognitive effort and suggested combining sentence-level QE scores with phrase or word-level QE indications. Taking on this direction, \newcite{shenoy-etal-2021-investigating} tried to determine a quality threshold at which QE is actually beginning to be useful for PE, and how to best present word-level QE information. They reported only limited productivity gains, arguing that QE systems need an F1 score of at least 80\% to support PE, an accuracy level that at the time had not been not reached. 

With the increased quality brought by neural MT and LLMs, errors became even harder to detect, which renewed interest in QE for PE. Using human ratings instead of automatic QE scores, \newcite{liu-etal-2025-introducing} found that sentence-level QE enhanced students' PE productivity only for high-quality segments. However, this varied by expertise level while inaccurate highlighting introduced confusion. 
The study most closely related to our work is that of \newcite{sarti2025qe4pewordlevelqualityestimation}, who systematically compared several error highlighting conditions in two language pairs (En$\rightarrow$It/Nl): a supervised state-of-the-art QE model trained on human annotations (xCOMET), an unsupervised method exploiting MT model uncertainty, oracle spans derived from consensus human post-edits, and a no-highlight baseline. They similarly reported limited or no overall productivity gains, even with oracle spans. 
Effectiveness varied considerably across language pairs, domains, and post-editors, pointing to translator attitudes, propensity to edit and working styles as confounding factors.

These findings collectively suggest that error highlighting alone is insufficient to reliably support post-editors. A complementary direction is to provide explicit translation suggestions as alternatives, either in a static ~\cite{10.1145/3173574.3174098} or interactive fashion~\cite{knowles2019user,alabau2016learning,briva2023impact}. This approach has been shown to increase productivity and help translators interpret suggestions more effectively. 
In this work, we contribute to this direction by assessing alternatives to QE-based error highlighting through APE-based error highlights and correction suggestions.

\section{Methodology}
\subsection{Data}
The data for the PE task comes from two domains: news and biomedical. Both domains contain interesting characteristics for MT. The biomedical domain was found to be rather challenging both for translation and QE models due to terminology and style~\cite{neves-etal-2024-findings}, and news texts required sensitivity to journalistic style, cultural references, and register
. Four 200-word English excerpts per domain were selected. The biomedical texts come from the QE4PE corpus \cite{sarti2025qe4pewordlevelqualityestimation} to allow for comparison with previous work. 
For the news domain, four texts were selected from the devsets of the WMT24 news shared task and 
adjusted to 200-word segments. 


\subsection{Models}
\textbf{Translation}: The selected excerpts were translated into Dutch (NL) using xTower-Instruct-13B-v0.1~\cite{treviso2024xtower} with the default template provided on HuggingFace\footnote{https://huggingface.co/Unbabel/TowerInstruct-13B-v0.1}. 

\textbf{Error spans}: xCOMET-XXL~\cite{Guerreiro24xCOMET} was used to automatically identify error spans ({\color{orange}H-QE}). 

\textbf{Translation corrections}: The texts and error spans obtained from xCOMET were put into xTower-Instruct-13B-v0.1 to generate error explanations and translation corrections (APE).

\subsection{Post-editing conditions}

\textbf{Regular post-editing ({\color{blue}PE})}: Translators post-edit without any highlights.

\textbf{PE with QE error highlights ({\color{orange}H-QE})}: Error spans obtained from xCOMET were used to highlight the text. Minor errors were highlighted in yellow, major in orange.

\textbf{PE with APE error highlights ({\color{magenta}H-APE})}:
Looking at the obtained error spans from xCOMET, we noticed that the highlights were often not very informative for human post-editing since they contained highlights that were not easily interpretable, e.g. single characters (see Table~\ref{tab:conditions}). For this reason, we used the translation corrections (APE) obtained from xTower to identify error spans. 
We compared the raw MT output and APE'd sentence using Levenshtein distance. 
The spans that differ between MT and APE were highlighted as errors, regardless of whether they were marked as an error by xCOMET. As with this approach there was no way to determine the error severity, all errors were highlighted as minor (yellow). Initial tests showed that the highlights obtained with this approach better correspond to human PE edits compared to QE spans (Appendix~\ref{app:overlap}).

\begin{figure*}[ht]
    \centering
    \includegraphics[width=\linewidth]{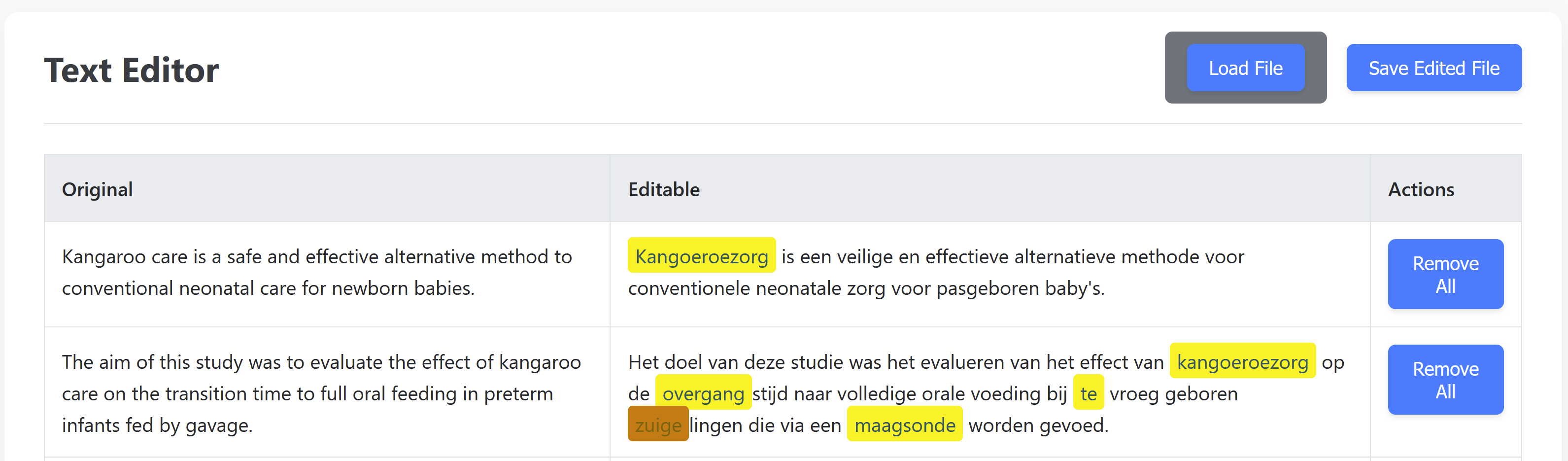}
    \caption{SmartPE: Post editing with error highlights ({\color{orange}H-QE} and {\color{magenta}H-APE}). Major errors in \colorbox{BurntOrange}{orange}, minor in \hl{yellow}.}
    \label{fig:interface-spans}
\end{figure*}

\begin{figure*}
    \centering
    \includegraphics[width=\linewidth]{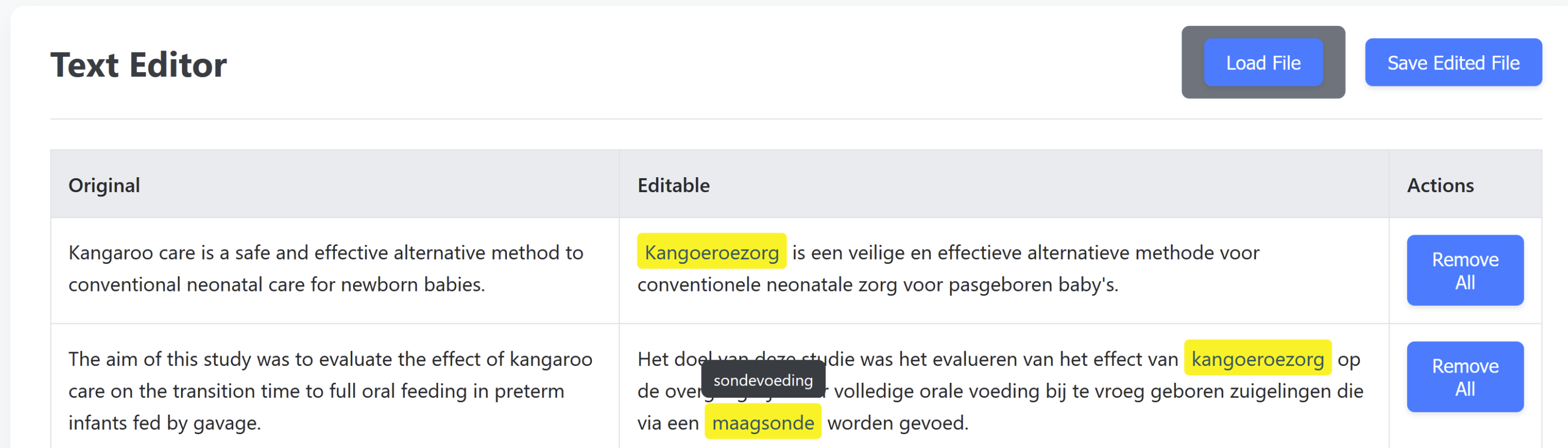}
    \caption{SmartPE: Post editing with error highlights and correction suggestions ({\color{SeaGreen}S-APE}).}
    \label{fig:interface-suggestions}
\end{figure*}

\textbf{PE with APE error highlights and correction suggestions ({\color{SeaGreen}S-APE})}: The error spans identified in {\color{magenta}H-APE} were 
paired to their APE correction.

To test the ability of translators to identify critical errors, two critical errors were manually inserted in each text (negation, serious mistranslation, serious omission) before annotating the errors. Out of the 16 total inserted critical errors, only 11 were annotated by xCOMET and 10 by xTower. However, since we wanted to determine whether highlights really help translators spot critical errors, we manually added major error tags around the missed errors. 

\subsection{Participants}
Eight professional translators, native Dutch speakers were recruited through a job post in professional groups and personal communication. Participants were full-time freelancers, experienced  translators in both domains (median: $>$10 years; 87.5\% with 5+ years) with moderate post-editing experience (median: 5-10 years). 

To assess the impact of each method on quality, all MT outputs and the eight post-edited versions were evaluated by one translator with 12 years of experience in medical and news texts using the Error Span Annotation (ESA) protocol~\cite{kocmi-etal-2024-error} and sentence-level direct assessment (DA) \cite{graham2017can}. 

\subsection{Post-editing task}
Before starting the task, the translators answered a questionnaire on their translation and PE experience, as well as perceptions on MT, PE, human translations, and their confidence in editing human/machine translations. All of them participated in a preparatory session to discuss the task and familiarise themselves with the interface. The post-editing guidelines can be found in Appendix~\ref{app:pe-guidelines}.

The PE task was conducted online at the translators' usual working space and equipment. Counterbalancing in a Latin square design (8 texts--2 per condition-- x 8 translators) was used to control for order effects and text difficulty. There was no time limit and the translators were instructed to take breaks between texts to avoid fatigue effects. The sessions were screen-recorded. After post-editing each text, the translators answered a short questionnaire on the perceived text difficulty, MT quality, error span quality and usefulness, perceived quality and usefulness of translation suggestions (where applicable). At the end of the PE task, each translator participated in a semi-structured interview to collect quantitative feedback on their user experience and confidence.

\subsection{Interface}
We developed \textbf{SmartPE}, a basic and simple-to-use interface for post-editing with highlights and correction suggestions in JavaScript. The interface logs the number of keystrokes and time spent editing a segment once it is active (clicked). Using this plain interface also helps control for translators' familiarity with CAT tools. Screenshots of the interface are shown in Figures~\ref{fig:interface-spans} and~\ref{fig:interface-suggestions}. The target text can contain highlights marking the type of error: light yellow for minor, orange for major. In addition, in the {\color{SeaGreen}S-APE} condition, correction suggestions appear in a black box once the user hovers the mouse over a highlight. By clicking on the suggestion, the highlighted text is substituted by the suggested text. Each interaction, including focus (which segment is active), keystroke, suggestion acceptance, and exit events (click outside the segment), is timestamped and logged to an in-memory activity log, which is exported as a CSV file upon saving. The interface is released open-source under MIT license.

\section{Results}
All results in Sections 4.1–4.5 are aggregated across the two domains (news and biomedical) to obtain comparisons for the primary research questions. Domain is examined separately in Section \ref{subsec:domain}. 
\subsection{Productivity} \label{sec:productivity}
\paragraph{Process data} To investigate whether the proposed features help translators post-edit texts more efficiently, we compare productivity, measured in characters/second, as the number of source characters processed over the text-level edit time, across the four conditions
. Text-level edit time corresponds to the timestamp 
of saving the translation minus the timestamp of opening the file. We opted for text-level completion time instead of segment-level times to account for time taken to read the text (without having activated a segment), searching outside the interface and other processes. 

\begin{figure}[ht]
    \centering
    \includegraphics[width=1\linewidth]{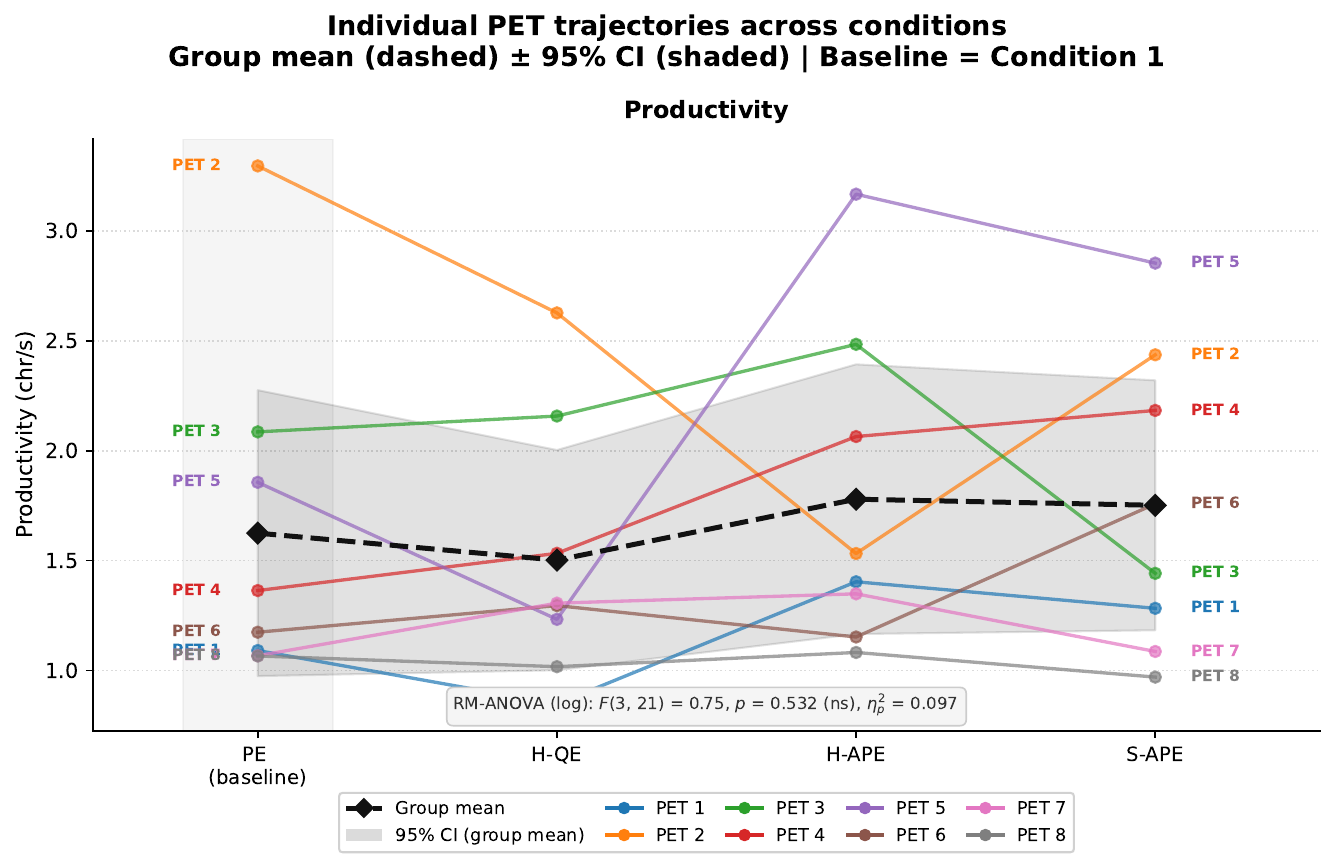}
    \caption{Productivity in characters/second for the eight post-editors (PET) and group mean (black interrupted line).}
    \label{fig:productivity}
\end{figure}

Figure~\ref{fig:productivity} shows the productivity per individual translation (PET) and as a group mean. The group mean is nearly flat across conditions, showing no productivity gains compared to regular {\color{blue}PE}. The results were confirmed statistically using one-way repeated measures ANOVA on log-transformed PET-level means (see Appendix, Table~\ref{tab:stats}), which revealed no significant effect of condition on productivity
. 
We observe that individual PET trajectories cross considerably, demonstrating strong individual differences, in line with previous research~\cite{terribile2024productivity,sarti2025qe4pewordlevelqualityestimation}. For PET 2, productivity considerably drops when using {\color{magenta}H-APE} highlights, contrary to PET 5, where APE highlights and suggestions lead to large productivity gains.  

\paragraph{Perceived effect on productivity} In the interview, the translators were asked to rate the perceived effect of highlights/suggestions on their productivity. For error highlights, only 2 of 8 translators reported working faster with the tool (PET 4 and 5). Most translators (3) reported reduced productivity, primarily because they had to spend additional time identifying and deleting incorrect highlights. The three other translators reported that their productivity remained about the same. For the {\color{SeaGreen}S-APE} suggestions, however,
six participants thought their productivity had increased.

Even though process data do not show any clear average improvements in productivity, the {\color{SeaGreen}S-APE} suggestions were beneficial for some translators and gave them the impression of improved productivity. These findings are in line with previous work which noted strong individual differences in productivity among translators \cite{sarti2025qe4pewordlevelqualityestimation}.

\subsection{Quality} \label{sec:quality}
\begin{figure}[ht]
    \centering
    \includegraphics[width=\linewidth]{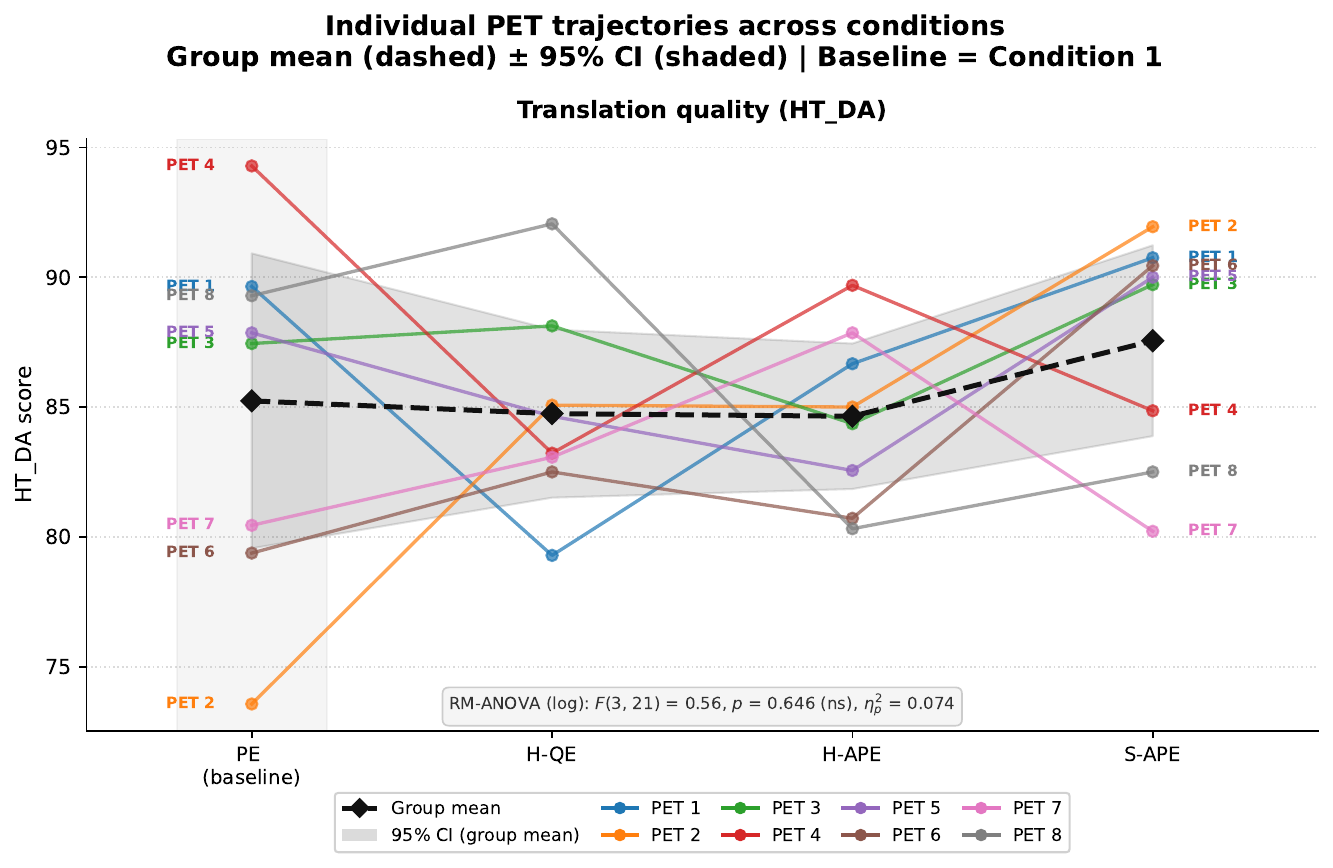}
    \caption{Final translation quality in terms of Direct Assessment scores per post-editor (PET) and group mean.}
    \label{fig:quality}
\end{figure}

\paragraph{DA ratings} We next explore the effect of PE condition on translation quality based on the experienced translator's DA ratings. Figure~\ref{fig:quality} shows that the quality of the final post-edited translations  was effectively the same across conditions, which is also confirmed by the RM-ANOVA analysis. As with productivity, we observe individual differences. A notable case is PET 2, for whom all forms of assisted PE led to quality improvements. Still, looking back at the productivity curve, this quality improvement could be due to spending more time on the tasks in general. Another interesting observation is that while the DA scores for the {\color{blue}PE} condition vary largely, DA scores cluster more strongly between 80-90 for the conditions relying on APE ({\color{magenta}H-APE} and {\color{SeaGreen}S-APE}), potentially consolidating quality levels among translators.

\paragraph{Perceived effect on quality} When asked if the error highlights helped improve the quality of the translation, half of the translators (4) thought that the quality did improve, while the rest stated that the highlights made no difference. For suggestions, almost all translators (7) found that the corrections improved the final quality, one of them stating \textit{translation quality improved a lot, by 60-70 percent, or maybe more}.

Even though no clear improvements in translation quality were observed, LLM suggestions still allow translators to achieve high quality translations (DA: 80-90), with {\color{SeaGreen}S-APE} suggestions leading to an improvement in perceived quality.

\subsection{Do highlights help translators spot and fix critical errors?}
\paragraph{Analysis of post-edits} Even though the overall quality was not significantly different among conditions, the question remains whether highlights and suggestions help translators spot critical errors. 
Table~\ref{tab:critical-accuracy} shows the percentage of critical errors (manually added before the PE task) that were fixed by the translators. Both {\color{orange}H-QE} and {\color{magenta}H-APE} perform equally (90\%), with only 3 out of the 128 critical errors missed respectively. This is an improvement compared to the simple {\color{blue}PE} condition, where merely 62\% of the inserted critical errors were fixed, showing that highlights have the potential to attract translators' attention to issues that may otherwise be missed. Seven critical errors were missed in the {\color{SeaGreen}S-APE} condition (84\% of errors fixed). Given that the highlights were the same between {\color{orange}H-QE} and {\color{SeaGreen}S-APE}, this minor drop may be simply due to chance. 

\begin{table}[ht]
    \centering
    \begin{tabular}{l|c|c|c|c}
     \toprule
         & {\color{blue}PE} & {\color{orange}H-QE} &{\color{magenta}H-APE} & {\color{SeaGreen}S-APE} \\\midrule 
         Number & 80 & 125 & 125 & 121 \\
    Perc. \%     & 62.5
    & 90.62 & 90.62 & 84.37 
    \\ \bottomrule
    \end{tabular}
    \caption{Number and percentage of critical errors fixed by condition.}
    \label{tab:critical-accuracy}
\end{table}

\paragraph{Perceived helpfulness in spotting errors} These results are contrary to what translators reported in their interviews. When asked whether highlights helped them detect errors  they may have otherwise overlooked, most participants (5) felt that the highlights did not make any difference, despite the large percentage of missed critical errors in the {\color{blue}PE} condition. 
On the contrary, most translators (7) mentioned that the translation corrections ({\color{SeaGreen}S-APE}) helped them detect errors, saying that the corrections proposed a better alternative.

\subsection{Are APE highlights more accurate and useful than QE highlights?}

\paragraph{Overlap of highlights with oracle edits}
To compare the accuracy of the spans obtained by QE and APE, we computed Average Precision (AP) and Area Under the Precision-Recall Curve (AUC) between automatic spans and error spans derived from human post-editing (oracle spans). As in \newcite{sarti2025qe4pewordlevelqualityestimation}, human oracle spans were derived by marking the spans that were edited by both translators in the {\color{blue}PE} condition.
Table~\ref{tab:ap-auc} shows that {\color{orange}H-QE} spans correspond more to human post-edited spans than {\color{magenta}H-APE}. This is contrary to the preliminary results observed using the QE4PE corpus in Table \ref{tab:ap-auc-qe4pe}, where {\color{magenta}H-APE} showed a higher agreement.

\begin{table}[ht]
\centering
\begin{small}
\begin{tabular}{lllll} \toprule
Method     &       AP & AUC \\ \midrule
{\color{orange}H-QE}         & 0.175                  & 0.707    \\
 {\color{magenta}H-APE}      & 0.094                  & 0.635    \\ \midrule
 Oracle (single transl.) & 0.51 & 0.73\\ \bottomrule    
\end{tabular}
\end{small}
\caption{Average precision (AP) and Area Under the Curve (AUC) for the spans obtained from different methods and human (oracle) post-edit spans. Oracle overlap per single translator corresponds to the average agreement between individual oracle post-editors and their consensus and is reported as an upper bound.}
\label{tab:ap-auc}
\end{table}

\paragraph{Perceived usefulness and accuracy}
The translators were blind to the method that was used to generate the highlights between {\color{orange}H-QE} and {\color{magenta}H-APE} conditions. To determine whether translators perceived any differences between the two highlight modalities, after post-editing each text with highlights they rated the usefulness and accuracy of error annotations on a 5-point Likert scale (\textit{Extremely accurate/useful} to \textit{Extremely inaccurate/useless}). Additionally, we computed the percentage of highlighted spans that were edited by translators, to determine upon which error spans translators were more inclined to act. The results are shown in Table~\ref{tab:perceived-spans}.

\begin{table}[ht]
\centering
\begin{small}
\begin{tabular}{lrrr} \toprule
& \textbf{\% High.ed.} & \textbf{Useful (1–5)} & \textbf{Accurate (1–5)} \\ \midrule
{\color{orange}H-QE}    & 48.33 \tiny{(15.99)} &     2.69 \tiny{(1.01)}    & 2.75  \tiny{(0.77)}          \\
{\color{magenta}H-APE} & 53.09 \tiny{(18.08)} &     3.38 \tiny{(0.96)}    & *3.50  \tiny{(0.82)}          \\
\bottomrule
\end{tabular}
\end{small}
\caption{Percentage of edited highlights (\% High.ed.) and self-reported rating of usefulness and accuracy of error highlights. Mean, and standard deviation in parentheses.}
\label{tab:perceived-spans}
\end{table}

In general, {\color{magenta}H-APE} highlights received higher usefulness and accuracy scores than {\color{orange}H-QE} highlights and were edited more often. {\color{magenta}H-APE} highlights were rated as slightly more useful ("somewhat useful") than {\color{orange}H-QE} highlights, which were considered between ``neutral" and ``somewhat not useful", even though this difference in perceived usefulness was not significant.
On the other hand, {\color{magenta}H-APE} highlights were found significantly more accurate (between ``not accurate nor inaccurate" and ``somewhat accurate") than {\color{orange}H-QE} highlights (closer to "somewhat inaccurate").\footnote{Usefulness: W = 8.0, p = .211\\ Accuracy: W = 2.5, p = .039 * } 

This tendency was confirmed by the interview responses, where translators were asked whether they observed any differences in the accuracy and usefulness of highlights between the texts they PEd in the different conditions. Five translators mentioned that they found the highlights in the {\color{magenta}H-APE} files more accurate, while three translators did not notice any clear differences. 

\subsection{How useful are correction suggestions?}
\paragraph{Percentage of accepted suggestions} We investigated the usefulness of correction suggestions ({\color{SeaGreen}S-APE}) by computing the percentage of suggestions that were accepted (inserted in the text) by the translators. Table~\ref{tab:correction} shows that about half the suggestions (49\%) were accepted on average by the translators. This does not mean that  half of the suggestions were wrong, since APE suggestions also included  preferential edits that translators did not consider necessary to implement. 
Still, the acceptance rates among translators vary widely (27-77\%). 

\paragraph{Perceived usefulness and accuracy} Based on the per-text ratings, translators rated the correction suggestions as ``somewhat useful" (mean 4.12). Accuracy ratings are slightly lower (mean 3.75) but still correction suggestions are considered ``somewhat accurate"
. In addition, we found an agreement between behaviour and perception, as translators who accepted more suggestions rated them as more useful as shown by a significant positive correlation between acceptance rate and usefulness at the text level\footnote{ ($\rho$ = .67, p = .005)}. The correlation with accuracy was non-significant.

\begin{table}[ht]
\centering
\begin{small}
\begin{tabular}{lrrr} \toprule
& \textbf{\% Sug.acc.} & \textbf{Useful (1–5)} & \textbf{Accurate (1–5)} \\ \midrule
{\color{SeaGreen}S-APE} & 48.95 \tiny{(19.56)}& 4.12 \tiny{(0.44)}        & 3.75 \tiny(0.76)          \\ \bottomrule
\end{tabular}
\end{small}
\caption{Percentage of correction suggestions accepted (\% Sug.acc.) and self-reported rating or usefulness and accuracy of correction suggestions. Mean and standard deviation in parentheses.}
\label{tab:correction}
\end{table}

Based on the interview responses, the correction suggestions ({\color{SeaGreen}S-APE}) were perceived more positively by the translators than the error highlights. Five translators stated the accuracy of correction suggestions was high and three of medium accuracy. One participant thought the accuracy was low, because the suggestion would not adapt to the rest of the sentence, as for example in interactive PE \cite{alabau2016learning}. 
Similarly, translators found the corrections useful (5) or mostly useful (2), explaining that the suggestion proposed a better alternative or increased the fluency of the text. 

\subsection{Are there differences between domains?} \label{subsec:domain}

\begin{figure}[ht]
    \centering
    \includegraphics[width=\linewidth]{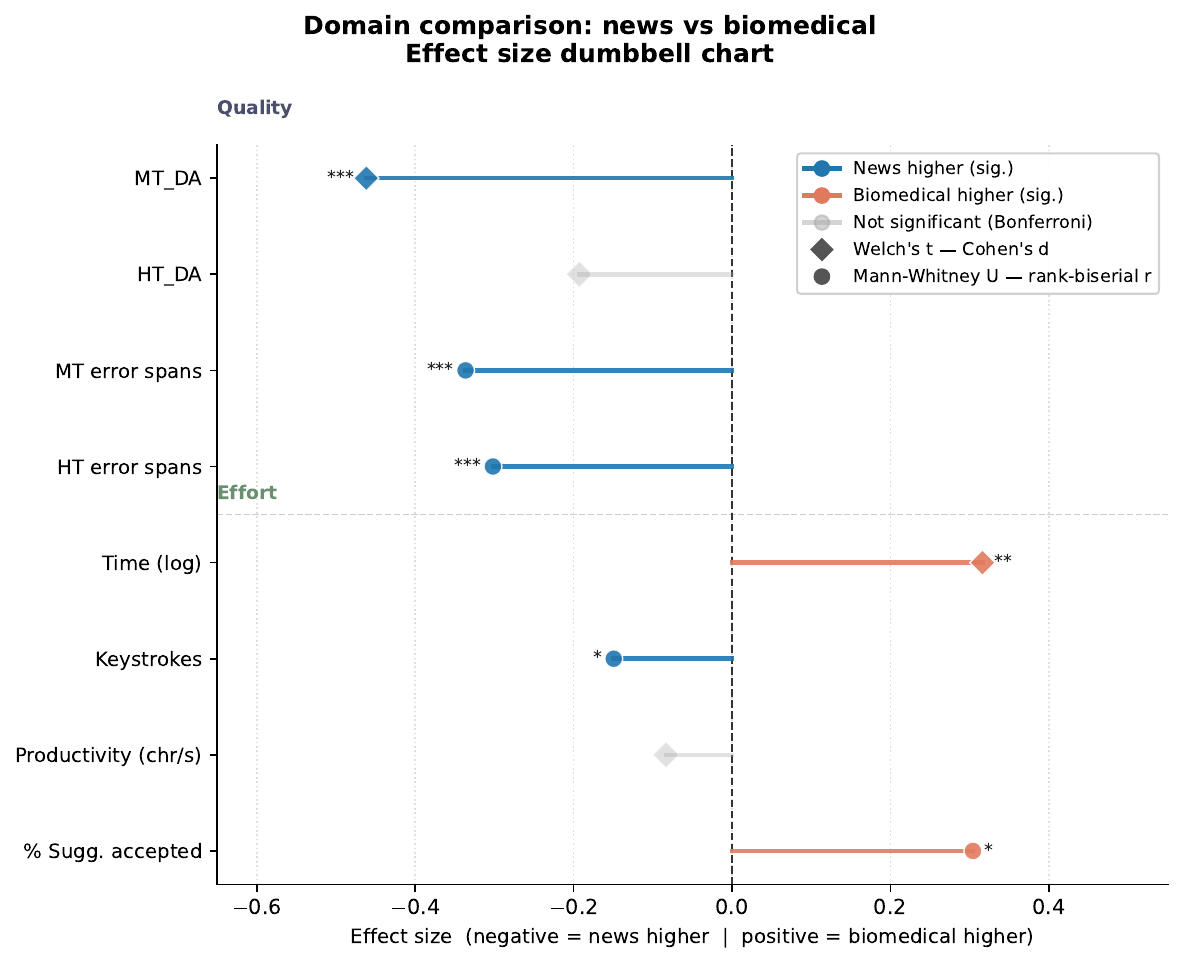}
    \caption{Differences in metrics between {\color{RoyalBlue}news} (left) and {\color{Bittersweet}biomedical} (right) domains.}
    \label{fig:domain}
\end{figure}

To examine whether post-editing behaviour and quality varied as a function of text domain, all variables were compared between news and biomedical texts (see Appendix~\ref{app:stats} for analysis). Results are shown in Figure~\ref{fig:domain}. The domain-level findings in terms of productivity and quality (HT\_DA) are largely consistent with those in Sections~\ref{sec:productivity} and ~\ref{sec:quality}, as productivity and quality did not show any statistically significant differences across domains (productivity 1.65 chr/s  vs 1.71 chr/s and DA scores 83.8 vs 87.2 for news and biomedical respectively). Despite this, news and biomedical texts showed differences across several dimensions of the post-editing process, with news texts proving more demanding in terms of both effort and quality. Looking at the sentence-level timings as recorded by SmartPE, translators spent more time post-editing sentences in news texts than in biomedical ones (69s vs 53s). However, given that the interface does not record time outside the active segment, this finding should be interpreted with care. 
Moreover, news texts required a higher number of keystrokes per sentence (54 vs 39) and translators had a tendency to accept less suggestions than for biomedical texts (37\% vs 60\%), showing that the suggestions may have been less well-suited to the more varied and idiomatic language of news content. In terms of MT quality, as assessed by the expert, news texts contained more errors overall (both major and minor) resulting in lower MT quality scores (MT\_DA 66 vs 71). This difference carried through to the post-edited output, where news texts retained a greater number of errors, even though overall post-edit quality scores (DA) did not differ significantly between domains. 


Subjective ratings of MT quality, error annotations, and translation suggestions did not differ significantly between domains, except for perceived text difficulty which was higher for biomedical texts (2.88 vs 3.47). This apparent paradox, where biomedical texts are rated as harder yet yield better MT and HT quality scores, may reflect the more constrained and technical nature of biomedical language. While translators find such texts cognitively demanding, the MT output is more predictable and requires less correction. \newcite{sarti2025qe4pewordlevelqualityestimation} found contrasting domain effects when comparing biomedical and social media texts: biomedical required time-consuming terminology verification, while social media edits were simpler and style-driven. Our results, even though not directly comparable, diverge from this pattern, as news texts proved more demanding than biomedical. These findings suggest that domain is a meaningful source of variation in post-editing effort and quality. 

\subsection{Translator perception and confidence}
\paragraph{Likes and dislikes} Thematic analysis of the interviews revealed that most participants found the error highlights distracting (3), disruptive to their workflow (2), or requiring extra manual effort (2). A small number of participants (2) liked working with highlights, mentioning that they provided \textit{an extra layer of support} or \textit{acted as a safety net}.  

Overall, the translation corrections were better received. Positive aspects were effort reduction (4), improving fluency of final translations (3) and increasing creativity by helping translators come up with translation solutions (3). Two translators felt that corrections gave more variety in wording choice. Other participants mentioned that the corrections were user-friendly (2) and easy to implement (3). 
This shows that not all forms of fine-grained feedback support translator efficiency equally, but solution-oriented assistance can have a positive effect on perceived effort even when error-focused annotations do not. 

\paragraph{Confidence} When asked whether the error highlights/correction suggestions made them feel more or less confident in their choices, the results were mixed. For error highlights, three translators thought their confidence had increased, while three translators were unsure whether their confidence had increased or decreased. This was particularly when they were dealing with news topics they were not as familiar with, as the error highlights made them question their knowledge. However, when using correction suggestions, the majority of translators (5) reported an increase in their confidence, attributing this to the corrections being accurate. Some mentioned corrections are \textit{a sort of backup system} or \textit{more like a sparring partner}. The rest of the participants (3) were either unsure or did not observe any difference in their confidence when using correction suggestions. Additionally, the effect on translators' editing choices seems to be stronger when using correction suggestions, since 5 translators thought that the error highlights influenced their editing choices compared to 7 for correction suggestions. 

In general, translators were willing to adopt LLM suggestions but stressed the importance of optionality (e.g. an option you can toggle on and off) and sustaining their agency in selecting which type of assistance they will be presented with. 

\section{Limitations and future work}
The present study is subject to several limitations that should be considered when interpreting its findings. First, the sample consisted of only eight professional translators, which severely limits statistical power and generalisability. While the within-subjects counterbalanced design partially compensates for the small sample size, the high degree of individual variation observed across conditions suggests that translator-specific factors, such as propensity to edit, prior attitudes towards MT, and individual editing styles, inevitably influenced both behavioural and quality outcomes. The within-subjects design also introduces the risk of learning and fatigue effects: with eight texts completed per translator, performance may have improved or deteriorated over the course of the session, although preparatory training was provided and translators were reminded to take breaks. Moreover, translation quality was assessed using DA scores produced by a single annotator, which limits the reliability of this measure
. While we believe that our design choices are justified within the context of our study, the impact on the outcomes cannot be entirely discounted.

Second, the study was conducted in a single high-resource language pair of  related languages (English$\rightarrow$Dutch). The same patterns may not hold for low-resource pairs where MT quality is substantially lower and APE tools are less mature or absent. Furthermore, we included two domains (news and biomedical)
and our results suggest that domain plays a significant role in determining the usefulness of support features (Section \ref{subsec:domain}). Although domain differences were reported, the limited number of texts means that text-specific difficulty cannot be fully disentangled from condition effects.  Lastly, although we attempted to mitigate individual, text and domain effects by ensuring a controlled evaluation setup for all translators and by using averaged judgments, we acknowledge that these effects may limit the reproducibility of our findings. 

Future work will employ more advanced statistical analyses based on mixed-effect models to account for individual differences among translators, such as prior MT attitudes and editing style, as covariates. In addition, further analysis will be conducted on the rich process data collected, including keystroke logs and screen recordings, which could reveal cognitive effort dynamics and attention patterns beyond the aggregate measures reported here. The multi-parallel dataset, combining automatic and human error annotations with eight PE versions, further enables span-level analysis of whether highlight accuracy predicts edit behaviour. Future work could also incorporate methods that avoid APE overcorrection in PE workflows~\cite{deoghare-etal-2025-giving}. 


\section{Ethical considerations and impact}
As LLMs introduce new applications that reshape translators' roles and transform established tasks such as post-editing, understanding how translators interact with LLM suggestions becomes increasingly important. This study tested LLM-based features for improving PE workflows, examining not only behavioural outcomes such as productivity, but also translation quality and translators' own perspectives on usefulness and the impact of these features on their workflows. Findings are in line with \newcite{sarti2025qe4pewordlevelqualityestimation}, who stressed that improved accuracy alone may not be sufficient to drive broader adoption of these techniques in post-editing workflows. 
Overall, translators showed a willingness to adopt LLM-generated suggestions, but consistently stressed the importance of optionality and the need to preserve their agency in selecting which type of assistance they are presented with. This aligns with broader critiques that translation technology has been driven by an AI emulation agenda rather than one of intelligence amplification or empowerment \cite{o2024human}. Our findings echo calls for more human-centred PE modalities \cite{briva2025human} and suggest that giving translators meaningful control over the type and degree of support rather than imposing it may be key to  adoption and preserving the ``dance of agency" \cite{olohan2011translators} between translators and their tools. By prioritising translators' perspectives alongside productivity gains, this work aims to contribute to methods that complement rather than replace human expertise. This is an important distinction at a time when such tools are frequently implemented and enforced without broad consensus from translators or adequate consideration of their professional needs.

All participants were professional translators who provided informed consent prior to participation. The research protocol ensured full anonymity and voluntary participation. In recognition of the time required for setup, familiarisation, and post-session interviews participants were remunerated at the higher end of standard rates for the Dutch translation industry. 
The materials developed for this study have been released publicly to promote transparency and reproducibility, enabling other researchers to build upon these findings.

\section{Conclusion}
Returning to the main question of this paper: do APE-based signals enable smarter edits by giving translators more accurate, actionable guidance about where and how to intervene? The answer is nuanced. 
Similar to previous studies \cite{shenoy-etal-2021-investigating,sarti2025qe4pewordlevelqualityestimation}, we found no significant behavioural evidence in favour of QE features in enhancing translators' productivity. While no support type lead to productivity gains, all conditions maintained a high level of translation quality. Between the two highlight modalities ({\color{orange}H-QE} and {\color{magenta}H-APE}) there is no clear winner: both modalities help translators detect critical errors and and prompt them to act upon highlights at comparable rates. Even though QE highlights show higher overlap with oracle edits (at least in this experiment), APE highlights obtain higher perceived accuracy and usefulness scores. 
Correction suggestions ({\color{SeaGreen}S-APE}) were better received than pure error highlights in subjective ratings and qualitative feedback and were found more informative. QE had a negative effect on translators' confidence, especially in cases where it was erroneous. However, when accompanied by a correction suggestion, erroneous highlights were reframed as preferential edits rather than distractions. These findings show that productivity, quality, and perception offer complementary lenses on what effective support looks like. Translators themselves provide the clearest answer: smarter editing suggestions are the ones that preserve translator agency through flexible, opt-in support rather than imposed assistance.


\paragraph{Acknowledgements} We kindly thank all the professional translators who took part in the experiment and evaluation. We also thank Ayşe Gül Açikgöz for  editing the interview transcripts.
This work was funded by the European Association for Machine Translation (EAMT) through its 2024 Sponsorship of Activities programme. 
The computational experiments were performed using the compute resources from the Academic Leiden Interdisciplinary Cluster Environment (ALICE) provided by Leiden University.

\bibliography{eamt26}
\bibliographystyle{eamt26}

\appendix
\section{Overlap between highlights and human edits}\label{app:overlap}
To determine which method for obtaining error spans overlaps more with  post-editing edits, before designing the user study we compared the automatically-predicted spans against spans derived from human post-editing. The data comes from the QE4PE dataset \cite{sarti2025qe4pewordlevelqualityestimation}) and consists of translations from English into Dutch in biomedical and social media domains. We evaluated the resulting spans using Average Precision (AP) and Area Under the Precision-Recall Curve (AUC) between the automatic spans and the error spans derived from human post-editing (oracle) at the word level. 

\begin{table}[ht]
\centering
\begin{small}
\begin{tabular}{lllll} \toprule
                   & \multicolumn{2}{c}{Biomedical}                   & \multicolumn{2}{c}{Social media}                 \\ \midrule
Method             & \multicolumn{1}{c}{AP} & \multicolumn{1}{c}{AUC} & \multicolumn{1}{c}{AP} & \multicolumn{1}{c}{AUC} \\ \midrule
{\color{orange}H-QE}         & 0.191                  & 0.531                   & 0.214                  & 0.581                   \\
 {\color{magenta}H-APE}      & 0.305                  & 0.614                   & 0.313                  & 0.674             \\ \midrule
 Oracle (single transl.) & 0.49 &  0.71  & 0.60 & 0.79 \\ \bottomrule    
\end{tabular}
\end{small}
\caption{Average precision (AP) and Area Under the Curve (AUC) for the spans obtained from different methods and human (oracle) post-edit spans.}
\label{tab:ap-auc-qe4pe}
\end{table}

The spans obtained from automatic post-edits based on translation corrections of xTower ({\color{magenta}H-APE}) demonstrated higher scores for both domains 
compared to those obtained from xCOMET ({\color{orange}H-QE}). This showed that APE suggestions align more closely with human post-edits than error spans identified by quality estimation methods and proved the motivation for the study. 

\section{Data}
The eight texts selected for the PE task come from the WMT2024 devsets (news) and the QE4PE dataset (biomedical) \cite{sarti2025qe4pewordlevelqualityestimation}, which were in turn extracted from PubMed from the WMT23 Biomedical Translation Task (Neves et al., 2023). The news texts were shortened to approx. 200 words. The texts selected are the following: 

\begin{table}[h]
    \centering
    \begin{tabular}{l|l} \toprule
     News    & Biomedical  \\ \midrule
     pa.52742    &  doc13 \\
     scotsman.87462 & doc18\\
     seattle\_times.799809 &  doc20 \\
     seattle\_times.800119 & doc34 \\ \bottomrule
    \end{tabular}
    \caption{Selected texts for the PE task.}
    \label{tab:placeholder}
\end{table}

\section{Statistical tests} \label{app:stats}
For the productivity and quality analysis, editing times and keystroke counts were log-transformed (log(x + 1) to handle zero values) prior to analysis to reduce skewness. The log transformation successfully normalised all variables. Shapiro-Wilk passes (p $> .05$) for every condition across all four outcomes. Sphericity also holds in all cases (Mauchly's p $> .05$)
. 
The analysis revealed no significant effect of condition on keystrokes, editing time, translation quality as measured by HT\_DA, or productivity.

\begin{table}[h]
\centering
\begin{small}
\begin{tabular}{llll} \toprule
\multicolumn{1}{c}{\textbf{Variable}} & \multicolumn{1}{c}{\textbf{F(3, 21)}} & \multicolumn{1}{c}{\textbf{p}} & \multicolumn{1}{c}{\textbf{partial $\eta$²}} \\ \midrule
Productivity   & 0.75  & .532 & .097 \\
Keystrokes   & 0.33      & .803   & .045    \\
Time & 0.14  & .937  & .019 \\
HT\_DA   & 0.56   & .646  & .074 \\
 \bottomrule                        
\end{tabular}
\end{small}
\caption{Results of Repeated Measures ANOVA reported in Sections~\ref{sec:productivity} and \ref{sec:quality}.}
\label{tab:stats}
\end{table}

Domain differences were examined using a combination of Welch's independent samples t-tests for continuous variables (MT\_DA, HT\_DA, productivity, and log-transformed time) and Mann-Whitney U tests for count data, proportions, and ordinal ratings (keystrokes, error spans, error counts, percentage of suggestions accepted, percentage of critical errors fixed, and subjective ratings). Effect sizes are reported as Cohen's d for Welch's t-tests and rank-biserial r for Mann-Whitney $U$ tests, with 95\% confidence intervals derived from 2,000 bootstrap samples. A Bonferroni correction was applied across all 13 sentence-level comparisons, yielding an adjusted significance threshold of $\alpha$ = .004.

\clearpage

\begin{table*}[ht]
\centering
\label{tab:domain}
\small
\begin{tabular}{llrrrrrc}
\toprule
& & \multicolumn{2}{c}{\textbf{News}} & \multicolumn{2}{c}{\textbf{Biomedical}} & & \\
\cmidrule(lr){3-4} \cmidrule(lr){5-6}
\textbf{Variable} & \textbf{Test} & \textit{M} & \textit{SD} & \textit{M} & \textit{SD} & \textbf{Effect} & \textbf{$p_{\text{bonf}}$} \\
\midrule
\multicolumn{8}{l}{\textit{Effort}} \\
\addlinespace[2pt]
Time (s)$^{\dagger}$    & Welch's \textit{t}  & 69.01 & 0.79 & 52.76 & 0.91 & $d$ = 0.32  & .011$^{*}$  \\
Keystrokes              & Mann-Whitney        & 54.23 & 67.83 & 39.56 & 44.53 & $r$ = $-$0.15 & .078        \\
Productivity (chr/s)    & Welch's \textit{t}  &  1.65 &  0.75 &  1.71 &  0.70 & $d$ = $-$0.08 & 1.000       \\
\% Sugg. accepted       & Mann-Whitney        & 37.42 & 37.81 & 60.45 & 41.59 & $r$ = 0.30  & .099        \\
\% Crit. errors fixed   & Mann-Whitney        &  0.79 &  0.31 &  0.83 &  0.30 & $r$ = 0.07  & 1.000       \\
\addlinespace[6pt]
\multicolumn{8}{l}{\textit{MT quality}} \\
\addlinespace[2pt]
MT\_DA                  & Welch's \textit{t}  & 60.66 & 23.49 & 71.73 & 24.56 & $d$ = $-$0.46 & $<$.001$^{***}$ \\
MT error spans          & Mann-Whitney        &  1.76 &  1.04 &  1.15 &  1.08 & $r$ = $-$0.34 & $<$.001$^{***}$ \\
MT major errors         & Mann-Whitney        &  0.62 &  0.67 &  0.39 &  0.55 & $r$ = $-$0.18 & .002$^{**}$ \\
MT minor errors         & Mann-Whitney        &  1.14 &  0.90 &  0.76 &  0.99 & $r$ = $-$0.26 & $<$.001$^{***}$ \\
\addlinespace[6pt]
\multicolumn{8}{l}{\textit{HT quality}} \\
\addlinespace[2pt]
HT\_DA                  & Welch's \textit{t}  & 83.78 & 17.82 & 87.20 & 17.70 & $d$ = $-$0.19 & .631        \\
HT error spans          & Mann-Whitney        &  1.23 &  1.10 &  0.68 &  0.85 & $r$ = $-$0.30 & $<$.001$^{***}$ \\
HT major errors         & Mann-Whitney        &  0.10 &  0.37 &  0.12 &  0.38 & $r$ = 0.02  & 1.000       \\
HT minor errors         & Mann-Whitney        &  1.13 &  1.03 &  0.56 &  0.79 & $r$ = $-$0.32 & $<$.001$^{***}$ \\
\addlinespace[6pt]
\multicolumn{8}{l}{\textit{Subjective ratings (text level, uncorrected)}} \\
\addlinespace[2pt]
Difficulty (1--5)       & Mann-Whitney        &  2.88 &  1.01 &  3.47 &  1.16 & $r$ = 0.30  & .031$^{*}$  \\
MT quality (1--4)       & Mann-Whitney        &  2.50 &  0.76 &  2.72 &  0.81 & $r$ = 0.20  & .141        \\
Ann. usefulness (1--5)  & Mann-Whitney        &  3.21 &  1.06 &  3.29 &  0.96 & $r$ = 0.04  & .811        \\
Ann. accuracy (1--4)    & Mann-Whitney        &  2.33 &  0.87 &  2.33 &  0.82 & $r$ = $-$0.01 & .936        \\
Sugg. usefulness (1--5) & Mann-Whitney        &  4.00 &  0.54 &  4.25 &  0.46 & $r$ = 0.22  & .367        \\
Sugg. accuracy (1--5)   & Mann-Whitney        &  3.88 &  0.64 &  3.63 &  0.92 & $r$ = $-$0.14 & .640        \\
\bottomrule
\end{tabular}
\caption{Domain comparison results: news vs biomedical texts. Welch's independent samples \textit{t}-tests were used for continuous variables; Mann-Whitney \textit{U} tests for counts, proportions, and ordinal ratings. Effect sizes are Cohen's \textit{d} (Welch's \textit{t}) and rank-biserial \textit{r} (Mann-Whitney \textit{U}). \textit{p}-values are Bonferroni-corrected across 13 sentence-level comparisons ($\alpha_{\text{adj}}$ = .004); rating-level variables are not corrected. Negative effect sizes indicate news texts score higher; positive values indicate biomedical texts score higher.}
\begin{tablenotes}
\small
\item $^{\dagger}$ Time values are geometric means (seconds); test performed on log-transformed values.
\item Significance (Bonferroni-corrected): $^{***}$ $p < .001$; $^{**}$ $p < .01$; $^{*}$ $p < .05$.
\item Subjective ratings are reported at text level ($N$ = 32 per domain for difficulty/MT quality; $N$ = 24 for annotation ratings; $N$ = 8 for suggestion ratings) and are not Bonferroni-corrected.
\end{tablenotes}
\end{table*}

\clearpage

\section{Post-editing guidelines}
\label{app:pe-guidelines}

This study investigates whether error annotations and translation correction suggestions provided by large language models can help professional translators post-edit machine-translated texts.

You are required to post-edit 8 short texts (200 words each-1600 words total) in biomedical and news domains under three conditions: 1) simple post-editing, 2) post-editing with error annotations and 3) post-editing with error annotations and suggestions. The task was conducted in an online interface. The task will be conducted in an online interface. After post-editing each file, you are asked to answer a few short questions about the translation. At the end of the task, we will conduct a short interview to collect feedback on your experience with post-editing.

\subsection{Interface-instructions of use}

\paragraph{1. Post-editing} Post-edit in the parallel interface as you would normally do in any CAT tool.

\paragraph{2. Post-editing with error annotations} Minor errors are highlighted in yellow and major errors in orange. Editing one of the highlighted words will make the highlight disappear. After finishing post-editing the sentence, click on \textbf{Remove all} to remove any remaining highlights.

\begin{figure}[ht]
    \centering
    \includegraphics[width=\linewidth]{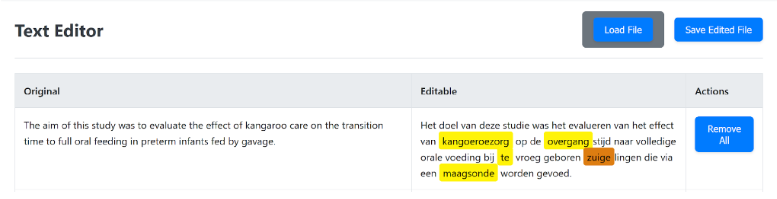}
    \caption{Example of the post-editing interface showing error annotations with minor errors highlighted in yellow and major errors in orange.}
    \label{fig:placeholder}
\end{figure}

\paragraph{3. Post-editing with error annotations and suggestions} Hovering the mouse over highlighted text will show a translation suggestion in a black box above the highlight. To adopt the suggestion, click on the black box. This will substitute the highlighted text with the translation suggestion. If you do not want to adopt the suggestion, continue post-editing as usual. After finishing post-editing the sentence, click on \textbf{Remove all} to remove any remaining highlights.

\begin{figure}[ht]
    \centering
    \includegraphics[width=\linewidth]{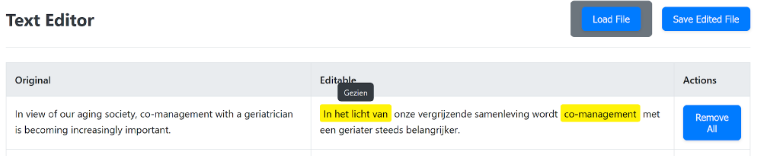}
    \caption{Post-editing interface showing error annotations with suggestions.}
    \label{fig:placeholder}
\end{figure}
\subsection{Task protocol}

\paragraph{Setting up your work environment}
\begin{itemize}
    \item Make sure you have a space where you can work without distractions.
\item Make sure to familiarise yourself with the interface before you start.
\item Join the Teams meeting. We will ask you to share your screen (only the interface window) and the meeting will be recorded. 
\end{itemize}

\paragraph{Workflow} \begin{itemize}
    \item Open the interface by double-clicking on the `main' file in the interface folder.
\item When prompted to enter your login code to start, type your participant code (e.g. PET1) and click on \textbf{Start editing}.
\item Click on \textbf{Load file} and select the file to post-edit.
When you have finished post-editing, click on \textbf{Save Edited File}. Two windows will open; one for saving a csv file, and a second one for saving a txt file. Make sure to save \textbf{both files}.
\item Close the interface.
\item Answer the short questions we provide you through Teams.
\item Take a break.
\end{itemize}

\paragraph{After finishing the task} 
\begin{itemize}
    \item Upload all the files (csv and txt) in the shared folder we have provided.
\item Notify us via Teams in order to start the interview.
\end{itemize}

\paragraph{Requirements} 
\begin{itemize}
    \item You are asked to post-edit each file by applying the minimal edits required to transform the MT into a correct translation of publishable quality. Follow your normal working processes, but do not spend excessive time on “polishing” any given wording or on researching information.
\item Focus on one task at a time.
\item Follow exactly the file order that was assigned to you.
\item Make sure to take a break of at least 5 mins between each file.
\item Please try not to take breaks when working on a file.
\item If you do need a break for any reason (toilet break, unforeseen interruptions etc.) please make sure to write down the start and finish time of the break, and do not close the interface window.
\item You may use any resources you need. You may use machine translation (e.g. Google, DeepL) only for inspiration, but do not copy entire MT’d segments into the interface.
\item The final deliverable should not present any highlights.

\end{itemize}

\end{document}